\documentclass[conference]{IEEEtran}
\IEEEoverridecommandlockouts
\usepackage{utfsym}
\usepackage{pifont}
\usepackage{stfloats}
\usepackage{url}
\usepackage{verbatim}
\usepackage{graphicx}
\usepackage{cite}
\usepackage{amsmath,amssymb,amsfonts}
\usepackage{algorithmic}
\usepackage{graphicx}
\usepackage{textcomp}
\usepackage{xcolor}
\usepackage{multirow}
\usepackage{booktabs}
\usepackage{bm}
\usepackage{xcolor}

\usepackage{makecell}
\def\BibTeX{{\rm B\kern-.05em{\sc i\kern-.025em b}\kern-.08em
    T\kern-.1667em\lower.7ex\hbox{E}\kern-.125emX}}
\begin{document}

\title{A New Method on Mask-Wearing Detection for Natural Population Based on Improved YOLOv4\\

\thanks{Xuecheng Wu, Mengmeng Tian, and Lanhang Zhai are all with school of Cyber Science and Engineering, Zhengzhou University. \emph{(Corresponding authors: Xuecheng Wu.)}
}
}
\author{\IEEEauthorblockN{Xuecheng Wu}
\IEEEauthorblockA{
\textit{Zhengzhou University}\\
Zhengzhou, Henan 450002, China \\
wuxc@stu.zzu.edu.cn}
\and
\IEEEauthorblockN{Mengmeng Tian}
\IEEEauthorblockA{
\textit{Zhengzhou University}\\
Zhengzhou, Henan  450002, China\\
tmm@stu.zzu.edu.cn}
\and
\IEEEauthorblockN{Lanhang Zhai}
\IEEEauthorblockA{\textit{Zhengzhou University}\\
Zhengzhou, Henan  450002, China \\
zhailhang@163.com}
}

\maketitle

\begin{abstract}
Recently, the domestic COVID-19 epidemic situation is serious, but in public places, some people do not wear masks or wear masks incorrectly, which  requires the relevant staff to instantly remind and supervise them to wear masks correctly. However, in the face of such an important and complicated work, it is very necessary to carry out automated  mask-wearing detection in public places. This paper proposes a new mask-wearing detection method based on improved YOLOv4. Specifically, firstly, we add the Coordinate Attention Module to the backbone to coordinate feature fusion and representation. Secondly, we conduct a series of network structural improvements to enhance the model performance and robustness. Thirdly, we adaptively deploy the K-means clustering algorithm to make the nine anchor boxes more suitable for our NPMD dataset. The experiments show that the improved YOLOv4 performs better, exceeding the baseline by 4.06\% AP with a comparable speed of 64.37 FPS.
\end{abstract}

\begin{IEEEkeywords}
YOLOv4, Coordinate Attention, K-means Clustering, Mask Wearing Detection. 
\end{IEEEkeywords}

\section{Introduction}
The new coronavirus can survive in various droplets for 24-48 hours. Wearing a mask and covering our mouth and nose in public is an effective approach to prevent infection. However, in some public places such as shopping malls and subway stations, some people do not take the initiative to wear masks or wear masks incorrectly. As a result, we must carry out tasks related to mask-wearing detection. In order to efficiently detect and reduce the impact of detection on the masses, it is necessary for us to apply computer vision technology to mask wearing detection. Existing detectors have improved the detection effectiveness to a certain extent, but they are susceptible to the effects of the external environment, the shape and color of the masks, and the detection accuracy rate of incorrect mask-wearing conditions is low. 

 In recent years, methods of applying neural networks to complete object detection tasks have emerged in an endless stream, and many object detection methods based on region proposals such as Faster R-CNN have been proposed. Later, related researchers proposed one-stage object detectors, and the YOLO series has also developed rapidly. The YOLO detector \cite{b4} was first proposed by Joseph Redmon et al. in 2015 and developed YOLOv2 \cite{b5} and YOLOv3 \cite{b7}. In 2020, Alexey Bochkovskiy et al. proposed YOLOv4 \cite{b1}, which is based on YOLOv3 and applies many emerging methods to achieve an optimal balance between speed and accuracy. In reality, the application of existing mainstream detection methods to mask wearing detection of the natural population will be affected by various factors, such as different styles and colors of masks, the skin color of the wearer, and weather. Therefore, the accuracy of the detectors is reduced to a certain extent, the robustness is poor, and the detectors can not meet the real-time requirements of natural scenes. Among them, Faster R-CNN has a high accuracy rate, but the detection speed can not meet the basic real-time requirements due to the limitation of its network structure; The performance of YOLOv3 \cite{b7} for small objects is not ideal, and the overall detection performance is also relatively terrible. 

Based on the abovementioned problems, this paper further optimizes the YOLOv4 \cite{b1} and deploys the Coordinate Attention Module to coordinately represent the inter-channel relationship and precise positional information for the intermediate feature maps. We then adjust the structure of the original network, significantly enhancing the depth and capacity of the overall network. Moreover, we utilize the K-means clustering algorithm to make the nine anchor boxes more suitable for our NPMD dataset. In this approach, the improved YOLOv4 proposed in this paper can better complete the task of mask wearing detection in natural scenarios.

\section{Methodology}

\begin{figure}[t]
\setlength{\belowcaptionskip}{-0.4cm}
\centering
\includegraphics[scale=0.458]{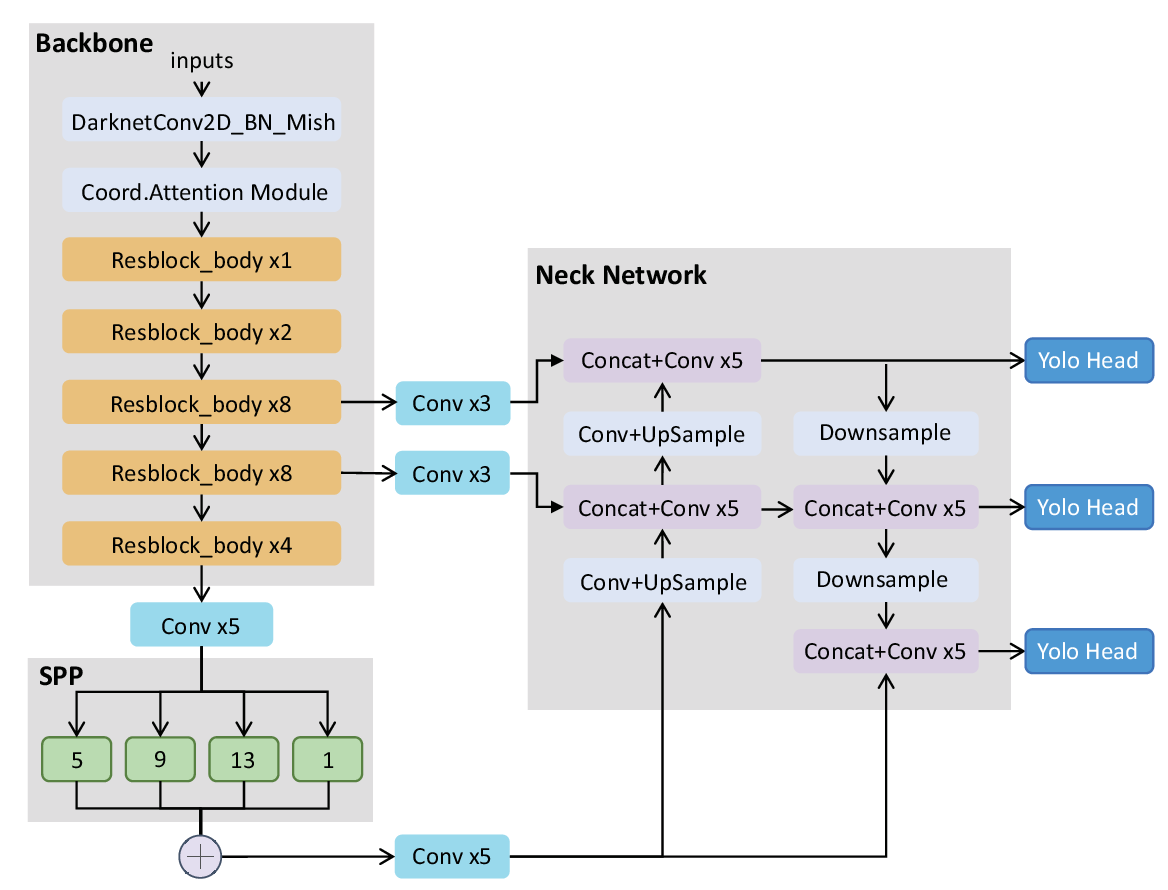}
\caption{The overall illustration of the improved YOLOv4 proposed in this paper. In this context, SPP denotes the layer of spatial pyramid pooling. }
\label{fig:1}
\end{figure}

The network structure of YOLOv4 is composed of a backbone, a neck network, and three YOLO Heads for different levels. In this paper, we specifically optimize the YOLOv4 for the characteristics of the mask wearing detection for the natural population. The overall illustration of our improved YOLOv4 is shown in Fig.~\ref{fig:1}.

\subsection{Coordinate Attention Module}
Coordinate Attention \cite{b2} is a new attention mechanism proposed by Qibin Hou and others at the National University of Singapore. This attention mechanism innovatively embeds specific positional information into inter-channel attention. This mechanism solves the common problems in SE, BAM and CBAM, which has better results and avoids introducing significant computational overhead. In this paper, after the convolution transformation layer denoted as DarknetConv2D-BN-Mish in the backbone, a Coordinate Attention Module is added before the residual convolution transformation blocks in order to strengthen the semantic representation of the shallow feature maps and obtain richer feature information over a larger region, which can further improve the performance of backbone. Compared with the original YOLOv4, the improved YOLOv4 combined with the Coordinate Attention Module can locate and identify the objects of interest more accurately. 

Coordinate Attention first performs maximum average pooling in the horizontal and vertical directions and then conducts transformation to encode the specific positional information accurately. Finally, the specific positional information is fused by weighting on the feature channels. The Coordinate Attention Module is divided into two steps: coordinate information embedding and coordinate attention generation.

In the step of coordinate information embedding, the Coordinate Attention Module first divides the global average pooling into a pair of 1D feature encoding operations so that the Coordinate Attention Module can capture the remote spatial interactions with precise positional information. Given the input \bm{$X$}, we deploy the two pooling kernels of size (H, 1) and (1, W) to encode each specific feature channel along with the horizontal and vertical directions. The abovementioned two transformations perform feature aggregation along with the horizontal and vertical directions and yield a pair of direction-aware feature maps. As a result, the output of the $c$-th channel at height $h$ and width $w$ can be formulated as Eq.~\ref{math:1} and Eq.~\ref{math:2}, respectively.

\begin{equation}
q_{c}^{h}(h)=\frac{1}{W} \sum_{0 \leq i<W} x_{c}(h, i)
\label{math:1}
\end{equation}

\begin{equation}
q_{c}^{w}(w)=\frac{1}{H} \sum_{0 \leq j<H} x_{c}(j, w)
\label{math:2}
\end{equation}

The $q_{c}^{h}$ and $q_{c}^{w}$ in Eq.~\ref{math:1} and Eq.~\ref{math:2} denote the outputs which are obtained after the average pooling operation along the horizontal and vertical directions, respectively.

In the coordinate attention generation, as shown in Eq.~\ref{math:3}, we first stack the specific pair of direction-aware feature maps generated in the step of coordinate information embedding and compress the number of feature channels by a $1 \times 1$ convolution transformation operation. We then encode the precise positional information in the horizontal and vertical directions through a BatchNorm layer and a ReLU layer. Afterward, $f$ is divided into two separate feature tensors $f^h$ and $f^w$ along the spatial dimensions, and two convolution transformation operations are utilized respectively to transform $f^h$ and $f^w$ to feature tensors with the same number of channels to the input \bm{$X$}. Then we can get $g^h$ and $g^w$ by normalized weighting, as shown in Eq.~\ref{math:4} and Eq.~\ref{math:5}.

\begin{equation}
f=\delta\left(F_{1}\left(\left[q^{h}, q^{w}\right]\right)\right)
\label{math:3}
\end{equation}
\begin{equation}
g^{h}=\sigma\left(F_{h}\left(f^{h}\right)\right)
\label{math:4}
\end{equation}
\begin{equation}
g^{w}=\sigma\left(F_{w}\left(f^{w}\right)\right)
\label{math:5}
\end{equation}

In Eq.~\ref{math:3}, $F_{1}(\cdot)$ denotes first stacking the two input feature tensors, followed by a $1 \times 1$ convolution transformation and a BatchNorm layer. $\delta(\cdot)$ represents the ReLU activation function and $f \in R ^ {C / r \times (H + W)}$ is the encoded intermediate feature map. In Eq.~\ref{math:4} and Eq.~\ref{math:5},  $F_{h}(\cdot)$ and  $F_{w}(\cdot)$ respectively denote a $1 \times 1$ convolution transformation operation.  $\sigma(\cdot)$ represents a sigmoid function.
The outputs $g^h$ and $g^w$ are then extended and utilized as the Coordinate Attention weights, respectively. The final output of Coordinate Attention is shown in Eq.~\ref{math:6}.

\begin{equation}
y_{c}(i, j)=x_{c}(i, j) \times g_{c}^{h}(i) \times g_{c}^{w}(j)
\label{math:6}
\end{equation}

In  Eq.~\ref{math:6}, $x_c(i, j)$ denotes the original input feature tensor, $g_{c}^{h}(i)$ and $g_{c}^{w}(j)$ denotes the horizontal attention weight and the vertical attention weight, respectively. And $y_{c}(i, j)$ denotes the weighted feature maps.

\subsection{Network Structural Improvement}
Although the neck network in the original YOLOv4 can strengthen the semantic representation capability of feature maps to a certain extent, it is susceptible to the natural scenes, mutual occlusion of people, and inconspicuous feature discrimination after people wearing masks during detection. Therefore, we make a series of neck network structural improvements to the original neck network. First, we increase the number of convolution transformation layers before and after the Spatial Pyramid Pooling layer from three to five. Second, inspired by the original YOLOv4, we increase the number of convolution transformation layers from one to three before $L3$ and $L4$ are fed into the neck network, respectively. The improved neck network deepens the capacity and depth of the overall network, obtaining larger receptive fields and richer semantic feature representations, significantly improving the model performance.

\subsection{K-means Clustering}
K-means clustering \cite{b3} is an iterative solution clustering analysis approach. The nine anchor boxes in the original YOLOv4 are computed initially by clustering the MS-COCO dataset utilizing K-means clustering. The MS-COCO contains 80 types of objects, and the sizes of the different objects varies greatly. If it is directly employed in the mask wearing detection, the sizes of some anchor boxes are unreasonable. Therefore, the K-means clustering is deployed again to perform clustering calculations on the NPMD dataset, which is collected by us, to obtain the sizes of the nine anchor boxes suitable for our dataset. In detail, first, we initialize the number of categories and cluster centers. We then calculate the distance, 1-IoU, between each bounding box and all cluster centers. After that, we choose the nearest cluster center as its category and utilize the average of each category cluster as the category center for the next iteration. We repeat the above two steps until the center position of each category does not change anymore. Finally, we can get the reasonable sizes of nine anchor boxes for our NPMD dataset.

\section{Experiments}
The experiments in this paper are based on the PyTorch1.8.1+cu101 framework, the programming language is python 3.7, the operating system is Ubuntu18.04, and the GPUs are NVIDIA RTX2080Ti*4, the integrated development environment is PyCharm 2020.3, and the input resolution of the overall network is $416 \times 416$. Moreover, we employ Adam as the optimizer for training. In the freezing training phase, we train a total of 50 epochs, and the initial learning rate is set to 0.001; In the global training phase, we train a total of 70 epochs, the initial learning rate is set to 0.0001, and the learning rate is adjusted utilizing the cosine annealing decay adjustment strategy.

\subsection{NPMD Dataset Introduction}\label{AA}
Currently, there are very few datasets on the detection of mask wearing in natural scenes, the detection environment is too single, and there is a general lack of images in the category of incorrect mask wearing. Therefore, we propose NPMD (Natural Population Mask Detection) dataset, which mainly includes images that meet our requirements crawled by the web crawler, those images selected from public datasets such as MAFA and RMFD which meet our requirements, and some images obtained by extracting frames from online videos. In addition, we have made some images by ourselves to expand the NPMD dataset further. The final original dataset contains 7,854 images in three categories: mask-wearing correctly, without mask-wearing, and mask-wearing incorrectly, involving multiple public natural scenes such as stations, subway stations, and supermarkets. Three common examples of mask-wearing incorrectly include uncovering the nose, uncovering the mouth and nose, and mask on the chin. The dataset is in the Pascal VOC format and annotated employing the LabelImg tool. Because the total amount of images in the original NPMD dataset is relatively small and the number of mask wearing incorrectly categories is also relatively small, we have adopted a variety of random geometric data augmentation, such as affine rotation transformation, Gaussian filtering, random color enhancement, and median blur. The random data enhancement method is utilized to further expand the total amount of images in NPMD dataset. After data enhancement, the total number of images in our NPMD is 11447. The number of each label in NPMD after random data augmentation is shown in Tab. \ref{tab:1}. In this approach, we can further reduce the risk of overfitting during the model training and improve the robustness and generalization of our proposed YOLOv4.

\begin{table}[]
\centering  
\caption{The number of labels by three categories in our proposed NPMD (Natural Population Mask Detection) dataset.}
\resizebox{6.2cm}{!}{
\begin{tabular}{c|c}
\toprule
\textbf{Category}      & \textbf{Number}\\ \midrule \midrule
Mask Wearing Correctly      & 14185                     \\ \midrule
Without Mask Wearing        & 7857                        \\ \midrule
Mask Wearing Incorrectly    & 6703                         \\ \midrule
Sum                         & \textbf{28745}                        \\ \bottomrule
\end{tabular}}
\label{tab:1}
\end{table}

\subsection{Evaluation Indicators}
The evaluation indicators chosen in our experiments are Average Precision ($AP$) and Mean Average Precision ($mAP$). The $AP$ measures the accuracy of model performance from the accuracy rate $P$ and the recall rate $R$. The accuracy rate represents the proportion of samples that are actually positive and predicted to be positive to all samples that are predicted to be positive. The recall rate represents the proportion of samples that are actually positive and predicted to be positive to all samples that are actually positive. The general formulas of $P$ and $R$ are shown as Eq.~\ref{math:7} and Eq.~\ref{math:8}, respectively.
\begin{equation}
P=\frac{TP}{TP + FP}
\label{math:7}
\end{equation}
\begin{equation}
R=\frac{TP}{TP + FN}
\label{math:8}
\end{equation}

In the abovementioned formulas, $TP$ represents a positive sample detected as correct; $FP$ represents a negative sample detected as a positive sample; $FN$ represents the positive sample detected as a negative sample.

$AP$ is calculated by the integral of the accuracy-recall rate curve. The higher the $AP$, the better the model performance, and its general formula is shown in Eq.~\ref{math:9}.
\begin{equation}
\mathrm{\emph{AP}}=\int P R d R
\label{math:9}
\end{equation}

$mAP$ is the average value of APs for each category, which is utilized to measure the average detection accuracy of multiple object categories. The value of $mAP$ can reflect the comprehensive performance of the detectors in all categories. The general formula of $mAP$ is shown as Eq.~\ref{math:10} below.
\begin{equation}
m A P=\frac{\sum_{i}^{c} C_{i}}{c}
\label{math:10}
\end{equation}

\subsection{Ablation Studies}
We propose a series of ablation experiments to verify the effectiveness of improved YOLOv4. We conduct four sets of ablation studies, and the specific schemes are as follows: the first group, the original YOLOv4, which is deployed as a control to determine whether our improvement component is effective or not; the second group is deployed to determine the effectiveness of the Coordinate Attention Module; the third group is used to determine the effectiveness of the neck network structural improvement; the last group is utilized to determine the effectiveness of K-means clustering for performance enhancement. The detailed results of ablation experiments are shown in Tab.~\ref{tab:2}.

Through the ablation experiments, we can clearly observe that each improvement component positively impacts the improved YOLOv4. The Coordinate Attention Module has significantly increased the AP by 2.67\%. Furthermore, we can conclude that the reason for the performance enhancement of improved YOLOv4 is that the Coordinate Attention Module is applied to the front-end of the backbone for guidance, resulting in a more assertive semantic representation of shallow-level feature images. At the same time, the improvement of the network structure further expands the receptive fields of deep-level feature images in disguise and then can extract more profound and richer feature information. The employment of K-means clustering makes the sizes of the nine anchor boxes we set more aligned with our NPMD dataset, further improving the model performance.

\subsection{Comparisons with other Advanced Detectors }
First of all, the APs of improved YOLOv4 for three categories are 96.12\%, 93.99\%, and 95.84\%, respectively. The mAP of our improved YOLOv4 achieves 95.32\%. Compared with the original YOLOV4, the mAP has increased by 4.06\% with a little bit of computational overhead.

In order to comprehensively evaluate the performance of our improved YOLOv4, we then compare the improved YOLOv4 with other CNN-based state-of-the-art object detectors on the NPMD dataset. The detailed results are shown in Tab.3. Specifically, ``Cat.1", ``Cat.2" and ``Cat.3" represent Mask Wearing Correctly, Without Mask Wearing, and Mask Wearing Incorrectly, respectively. We can see that compared with Faster R-CNN, SSD, YOLOv3 and the original YOLOv4, the improved YOLOv4 has significantly improved the APs of the three categories, and the mAP is also better than the abovementioned four SOTA detectors. Our improved YOLOv4 further improves the detection ability of small targeted objects and the processing of detailed features. The effectiveness of our proposed improved YOLOv4 is further confirmed by comparing it with the current mainstream object detectors.

\section{Conclusions}
The current situation of COVID-19 is severe and complex, and mask wearing is one of the most effective prevention methods. This paper proposes a new mask wearing detection method based on improved YOLOv4. The extensive experimental results on our NPMD dataset show that the improved YOLOv4 has good accuracy, can better meet the actual needs of epidemic prevention and control, and realize comprehensive and accurate mask wearing detection in natural scenes. 

\begin{table}[]
\centering  
\caption{The ablation experiment on the effectiveness of each improvement component in our improved YOLOv4. Moreover, we evaluate the model performances in terms of mAP(\%) on the NPMD dataset.}
\resizebox{8.5cm}{!}{
\begin{tabular}{c|c|c|c}
\toprule
\textbf{Method}      & \textbf{$\bm{\mathrm{mAP}}$(\%)} & \textbf{Parameters} & \textbf{FPS}\\ \midrule \midrule
Original YOLOv4      & 91.26              & 64.43M             & 69.26              \\ \midrule
+ CA Module         & 93.93               & 64.72M              & 66.82             \\ \midrule
+ Structural Improvement & 94.71          & 65.48M              & 64.37              \\ \midrule
+ K-means Clustering    & \textbf{95.32}            & 65.48M             & 64.37              \\ \bottomrule
\end{tabular}}
\label{tab:2}
\end{table}

\begin{table}[]
\centering 
\caption{Performance comparison of different mainstream object detectors on NPMD dataset. Moreover, we train all the models with the same learning schedule and optimizing parameters for a fair comparison.}
\resizebox{8.5cm}{!}{
\begin{tabular}{l|lll|l}
\toprule
\multirow{2}{*}{~~~\textbf{Method}} & \multicolumn{3}{l|}{\textbf{~~~~~~~~~~AP(\%)}}                              & \multirow{2}{*}{\textbf{mAP(\%)}} \\ \cline{2-4}
                   & \multicolumn{1}{l|}{\textbf{Cat.1}} & \multicolumn{1}{l|}{\textbf{Cat.2}} & \textbf{Cat.3} &                    \\ \midrule \midrule
Faster R-CNN                  & \multicolumn{1}{l|}{88.13} & \multicolumn{1}{l|}{89.14} & 92.11 & ~~89.79                  \\ \midrule
SSD \cite{b6}                 & \multicolumn{1}{l|}{84.72} & \multicolumn{1}{l|}{81.69} & 94.09 & ~~86.84                  \\ \midrule
YOLOv3 \cite{b7}                 & \multicolumn{1}{l|}{87.07} & \multicolumn{1}{l|}{89.27} & 93.74 & ~~90.02                  \\ \midrule
YOLOv4 \cite{b1}                  & \multicolumn{1}{l|}{90.29} & \multicolumn{1}{l|}{89.65} & 93.85 & ~~91.26                  \\ \midrule
\textbf{Ours-YOLOv4}                  & \multicolumn{1}{l|}{96.12} & \multicolumn{1}{l|}{93.99} & 95.84 & \textbf{~~95.32}                 \\ \bottomrule
\end{tabular}}
\label{tab:3}
\end{table}

\end{document}